\documentclass[11pt,a4paper]{article}
\usepackage[hyperref]{tto2019}
\usepackage{times}
\usepackage{latexsym}
\usepackage{booktabs} %
\usepackage[normalem]{ulem}
\usepackage{url}
\usepackage{xcolor,colortbl} %
\usepackage{algorithm}
\usepackage{algpseudocode}
\usepackage{amsmath}
\usepackage{amssymb}
\usepackage{enumerate}
\usepackage[shortlabels]{enumitem}
\usepackage[mathscr]{eucal}
\usepackage{graphicx}
\usepackage{subfigure}
\usepackage{makecell}
\usepackage{mathtools}
\usepackage{multirow}
\usepackage[font=rm]{caption}
\DeclareCaptionType{copyrightbox}
\usepackage{appendix}
\usepackage{shortvrb}
\usepackage{tabularx}
\usepackage{verbatim}
\usepackage{xspace}
\usepackage{listings}
\usepackage{textcomp}
\usepackage{fontawesome}
\usepackage[multiple]{footmisc}
\usepackage[all]{nowidow}
\usepackage{balance}

\makeatletter
\g@addto@macro{\UrlBreaks}{\UrlOrds}
\makeatother

\newcommand\LQ[1]{#1}
\newcommand\GW[1]{#1}

\newcommand{\pheadA}[1] {\vspace{1.2mm}\noindent\textbf{#1.}} %
\newcommand{\pheadB}[1]{\vspace{0.2\baselineskip}\noindent{\textbf{#1}}.~}%

\newcommand{\B}{\textbf}

\newcommand{\Patk}[1]{\mbox{\Pat@$#1$}}

\newcommand{\squishlist}{
 \begin{list}{$\bullet$}
  { \setlength{\itemsep}{0pt}
     \setlength{\parsep}{1pt}
     \setlength{\topsep}{1pt}
     \setlength{\partopsep}{0pt}
     \setlength{\leftmargin}{1.5em}
     \setlength{\labelwidth}{1em}
     \setlength{\labelsep}{0.5em} } }
 \newcommand{\squishend}{\end{list}}

\aclfinalcopy %

\newcommand{\hnda}{$\rm{NEW_A}$}
\newcommand{\hnds}{$\rm{NEW_S}$}
\newcommand{\ira}{$\rm{TWE}$}
\newcommand{\spa}{$\rm{SPE_A}$}
\newcommand{\sps}{$\rm{SPE_S}$}
\newcommand{\lstmr}{$\rm{LSTMR}$}
\newcommand{\lstm}{$\rm{LSTM}$}
\newcommand{\lr}{$\rm{LR}$}
\newcommand{\svm}{$\rm{SVM}$}
\newcommand{\wc}{$\rm{WC}$}
\newcommand{\tiw}{$\rm{TI}$-$\rm{W}$}
\newcommand{\tig}{$\rm{TI}$-$\rm{G}$}
\newcommand{\liwc}{$\rm{LIWC}$}
\newcommand{\all}{$\rm{ALL}$}
\newcommand{\thr}{$\rm{THR}$}

\definecolor{cadetgrey}{rgb}{0.66, 0.66, 0.66}

\title{Cross-Domain Learning for\\Classifying Propaganda in Online Contents}

\author{Liqiang Wang$^{1,2}$, Xiaoyu Shen$^1$\thanks{Now at Amazon, work done before joining}, Gerard de Melo$^3$, Gerhard Weikum$^1$\\
  Max Planck Institute for Informatic, Saarbr\"ucken, Germany$^1$ \\
  Shandong University, Jinan, China$^2$ \\
  Hasso Plattner Institute, University of Potsdam, Potsdam, Germany$^3$\\
  \texttt{\{lwang, xshen, weikum\}@mpi-inf.mpg.de, gdm@demelo.org} \\
  }
\date{}

\begin{document}
\maketitle
\begin{abstract}
 As news and social media exhibit an increasing amount of manipulative polarized content, detecting such propaganda has received attention as a new task for content analysis. Prior work has focused 
on supervised learning with training data from the same domain. However, as propaganda can be subtle and
keeps evolving, manual identification and proper labeling are very demanding. As a consequence, training data is a major bottleneck. 

In this paper, we tackle this bottleneck and present an approach to leverage cross-domain learning, based on labeled documents and sentences from news and tweets, as well as political speeches with a clear difference in their degrees of being propagandistic. We devise informative features and build various classifiers for propaganda labeling, using cross-domain learning. 
Our experiments demonstrate the usefulness of this approach, and identify
difficulties and limitations in various configurations of sources and targets for
the transfer step.
We further analyze the influence of various features, and characterize
salient indicators of propaganda.
\end{abstract}

\section{Introduction}

\subsection{Motivation and Problem}
\GW{Propaganda can be loosely defined as 
{\em ``misleading information that is spread deliberately to deceive and manipulate its recipients''} (see, e.g., \citealp{jowett2018propaganda} 
and {\href{https://www.britannica.com/topic/propaganda}{www.britannica.com/topic/propaganda}}).
}%
Various factors
of propaganda have been studied in the humanities, 
including emotionality of language, biased selection of information and deviation from facts, manipulation of cognition, and more \cite{ellul1973propaganda, silverstein1987toward,jowett2018propaganda}.
However, there is no consensus on the decisive
factors that tell whether a given article or speech is propagandistic 
or not.
In the modern digital world, the influence of propaganda on society
has drastically increased. 
Hence, there is also a major increase in computer science, computational linguistics
and computational sociology research on analyzing, characterizing
and, ultimately, automatically detecting propaganda \cite{DBLP:conf/ijcai/MartinoCBYPN20}.

To a first degree, one may think of propaganda as a variation of
fake news, and 
some works investigate propaganda as a refined
type of disinformation (see, e.g., \citealp{rashkin2017truth,wang2019understanding,shu2017fake}).
\GW{While false claims can be an element of propaganda, we think that fake news is merely
the tip of the iceberg, and that the persuasive and manipulative nature of propagandistic contents requires deeper approaches.}
Classifiers for propaganda detection need to better capture how propaganda is expressed in subtle ways by language style
and rhetoric or even demagogic wording. This holds for news as well
as social media posts and speeches.
In all these cases, correct information may be presented in
incomplete form or placed in distorted contexts, along with
manipulative phrases, in order to mislead the audience.

Prior work has mostly looked into news articles (e.g., \citealp{saleh2019team,barron2019proppy,da2019fine})
and tweets, and has typically focused on 
strongly polarized topics like the 2016 US election
and the related Russian Internet Research Agency (IRA) affair,
the UK Brexit discussion,
or political extremism.
\LQ{All these approaches consider propaganda detection
as a classification task assuming sufficient amounts of labeled in-domain training data.}
\LQ{For example, in the ``Hack the News'' datathon challenge\footnote{\url{https://www.datasciencesociety.net/hack-news-datathon/}\label{hackd}},
a large number of news articles \cite{barron2019proppy} and sentences \cite{da2019fine} from such articles
were annotated by distant supervision and human judgment, respectively, to train a variety of machine learning methods.}
The resulting F1 scores on the leaderboard of this benchmark
are amazingly high, around 90\%.
This may give the impression that propaganda detection is a solved
problem. However, most of the positively labeled samples are simple
cases of ``loaded language'' with strong linguistic cues independent of the topic.
Moreover, the learned classifiers
benefit from ample training data, which is
all but self-guaranteed in general.

In this paper, we question these prior assumptions, hypothesizing 
that propagandistic sources and speakers are sophisticated and
creative and will find new forms of deception
evading the trained classifiers.
\GW{The overall approach is still text classification; the novelty of our approach lies in {\em cross-domain learning}, where
domains denote different kinds of sources,
such as news articles vs.\ social media posts
vs.\ public speeches.}
We acknowledge that there is often a shortage of
perfectly fitting labeled data, and instead tap into
alternative sources that require a transfer step.
Specifically, we consider speeches and tweets, in addition
to news articles, at both article and sentence levels.

\subsection{Approach and Contribution}
Our goal is to build more general propaganda detectors, which can
leverage different kinds of data sources.
In particular, we tap on political speeches of notorious 
propagandists, such as Joseph Goebbels
(the Nazi's Reich Minister of Propaganda).
As it is very difficult (and often subjective) to label speeches
and their sentences in a binary manner, 
we pursue a pairwise ordinal approach where training data merely
ranks samples of a strongly propagandistic speaker against those
of a relatively temperate speaker.
We investigate to what extent models learned from such data
can be transferred to classifying news and tweets,
and we also study the inverse direction of learning from
news and tweets to cope with speeches.

\begin{figure}[ht!]
    \centering
    \includegraphics[width=1\columnwidth]{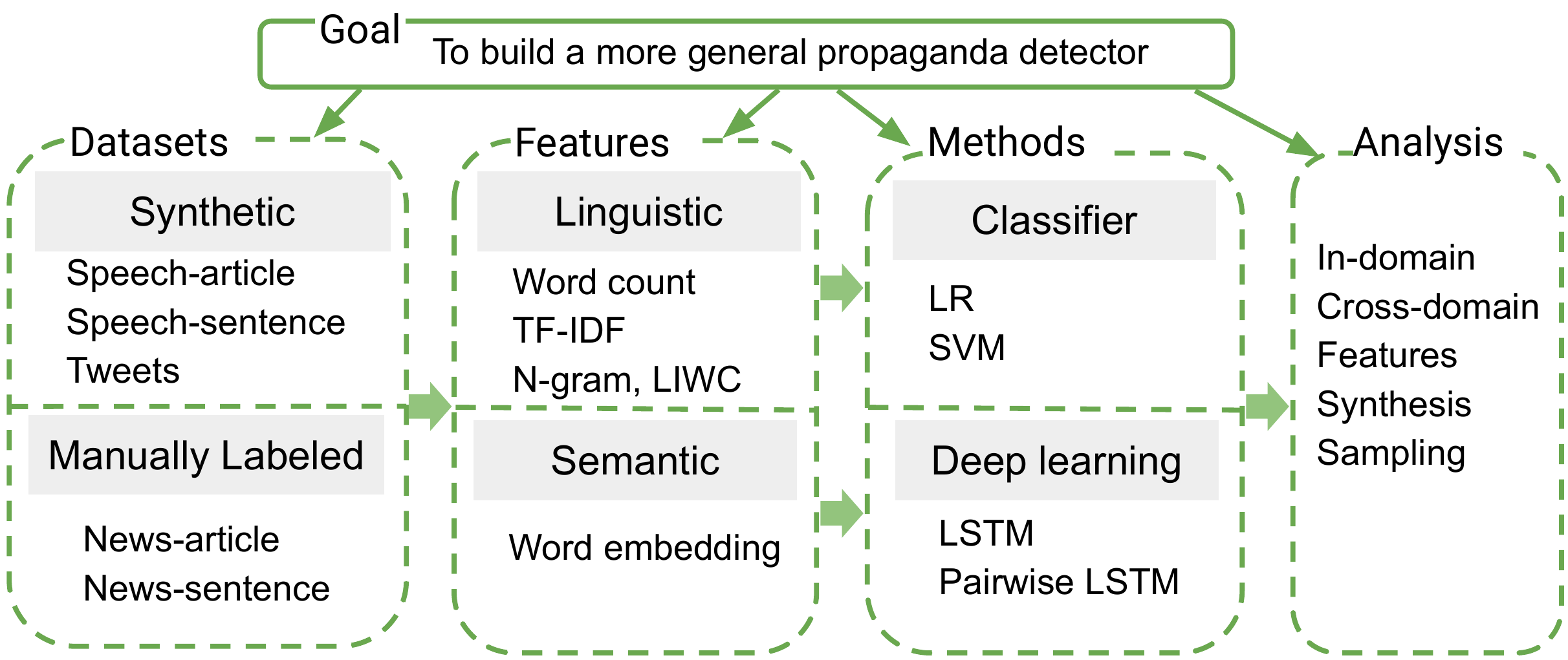}
	\caption{Framework for propaganda detection and analysis}
	\label{framework} 
\end{figure}

Figure~\ref{framework} illustrates our framework towards 
generalizable propaganda detection that overcomes the bottleneck
of directly applicable training labels and instead leverages
cross-domain learning.

\noindent The salient contributions of this paper are as follows.
\begin{enumerate}
    \item We introduce a framework for cross-domain learning of propaganda 
    classification. This comprises data collection, feature selection, training
    procedures, different learning methods, and analysis.
    \item We devise a pairwise ranking LSTM model, called \lstmr, 
    that uses a judiciously designed loss function to enhance cross-domain performance.
    \item We present experiments and analyses 
    that provide insights into the advantages as well as limitations of
    cross-domain learning for propaganda classification.
    \item All our datasets and code have been made publicly available 
    to support further research on understanding and countering
    advanced forms of propaganda.\footnote{\url{https://github.com/leereak/propaganda-detection}}
\end{enumerate}

\section{Related Work}
\pheadA{Propaganda Detection}
We summarize the state-of-the-art 
according to the type of underlying data sources:
the news, tweets or data to be fact-checked. 
\citet{saleh2019team} detect propagandistic content 
using a variety of well-engineered features. 
The study is based on a dataset of news articles
annotated for the SemEval 2019 task on ``Hyperpartisan News Detection''.
\citet{barron2019proppy} generated training and test data
by collecting news from blacklisted sites 
as propaganda, while considering the Gigaword News corpus as trusted news. 
\LQ{Further research related to propaganda detection on news articles includes work on multi-class labeling \cite{da2019fine} and on the related topic of hyperpartisanship \cite{potthast2018stylometric}.}

For Twitter as a data source, several datasets are widely used,
most notably the tweets on the Russia-based IRA \cite{miller2019topics,farkas2018ira,badawy2019characterizing}, contents spread by trolls or bots \cite{caldarelli2019role,williamson2019trends}, and extremist tweets \cite{nizzoli2019extremist,johnston2017identifying}.
Most of these studies rely on classifiers developed for bot detection to filter the propagandistic contents.
Deep learning methods are also used to detect propaganda \cite{johnston2017identifying,nizzoli2019extremist}.
As more emphasis is on the analytics, the above detection techniques are still in the formative stage, \LQ{by applying basic approaches like logistic regression, support vector machines, LSTMs, etc. These methods motivate us with regard to our feature engineering.} 
As mentioned above, some works take propaganda as a fine-grained type of fake news \cite{rashkin2017truth}.
Some approaches \cite{rashkin2017truth,wang2019understanding,shu2017fake,shu2019beyond}  
considered fact-checked statements from  sources such as PolitiFact, Snopes, BuzzFeed, etc. This data encompasses trusted content, satire, hoaxes, as well as propaganda, but the approaches do not develop propaganda-specific techniques.
In our view, sophisticated forms of propaganda are quite different
from hoaxes or plainly wrong statements, and call for custom-tailored
approaches.

\pheadB{Propaganda Analysis}
Analyses of propagandistic content have mostly looked into
quantifying its influence and spreading across networks.
\citet{timothy2017does} studied how propaganda influences public opinion by means of simulations.
Some studies \cite{caldarelli2019role,williamson2019trends} analyzed the role of bots in spreading propagandistic posts.
\citet{gorrell2019partisanship} examined
strategies of troll accounts. 
\citet{farkas2018ira} reported findings on propaganda in IRA tweets. 
The observations suggest that techniques are customized to
the targeted political agenda.
As for feature analysis, the work of
\citet{barron2019proppy} and \citet{potthast2018stylometric} investigated the effectiveness of linguistic and stylistic features. 
\citet{bisgin2019analyzing} analyzed extremism propaganda contents
in terms of entities, topics and targets.

Fake news analysis techniques \cite{resende2019mis,volkova2018misleading,shen2017estimation,wang2019understanding,zhang2018structured} can as well serve as inspiration to better understand propaganda. 
In particular, \citet{budak2019happened}
studied the prevalence and focus of fake news 
based on related tweets, news and interviews.
\citet{yang2019xfake}
used visualization techniques for analyzing fake news to make the detector more explainable.

\section{Datasets}
For our study, we compile five datasets from three domains with the aim of exploring cross-domain characteristics and performance of propaganda detection.

\pheadA{Speeches} We collected transcripts of speeches from 
four politicians, organized as ordered pairs.
\LQ{Trump\footnote{\url{https://factba.se}} and Obama\footnote{\url{http://obamaspeeches.com}} are considered as contemporary speakers,
largely talking about the same or related topics.
We consider the former as more propagandistic than the latter.
We use Joseph Goebbels (the Nazis' Minister of Propaganda, human-translated to English by the data provider)\footnote{\url{https://research.calvin.edu/german-propaganda-archive/goebmain.htm}}
and Winston Churchill (the Prime Minister of the UK)\footnote{\url{https://winstonchurchill.org/resources/speeches/}}
as prominent figures from the World War II era,
with the former being more propagandistic than the latter.}
\GW{We realize that all four of these politicians have given some propagandistic speeches. Our assumption is that, collectively and relatively, two of the speakers exhibit substantially less propaganda than the other two.}
The data is organized at two different levels of granularity: 
articles (\spa) and sentences (\sps).

\pheadA{News} The news dataset comes from the Hack the News Datathon\textsuperscript{\ref {hackd}}.
We combined and reorganized different datasets \cite{barron2019proppy,da2019fine} from this source,
to construct an article-level news corpus (\hnda) and a sentence-level
corpus (\hnds), both comprehensively annotated with binary labels:
propagandistic or normal.
Note that the articles for the sentence-level corpus are
completely disjoint from the ones in the article-level corpus.
So learning on one and testing on the other entails a challenging
cross-domain transfer as well.

\pheadA{Tweets} We combine two pre-existing collections of tweets
to construct this \ira\ dataset. \LQ{We consider the Twitter IRA corpus \cite{sean2017Testimony} with time period of 2016 as propagandistic, and the ``twitter7'' data from SNAP \cite{yang2011patterns} in 2009 as regular.}
\LQ{As ``twitter7'' encompasses around 476 million tweets, we under-sampled 8,963 tweets to maintain a balance with the IRA data. The tweets are randomly sampled from the June 2009 collection within the dataset.}
As with speeches, the data is cast into ordered pairs rather than
using absolute labels as ground truth.

As some of the data initially comes with a strong label skew
(with way more negative than positive samples), 
we apply under-sampling to construct corpora with
balanced positive and negative samples.
Table~\ref{dataset} summarizes our five datasets. 

\begin{table}[!ht]
	\centering
    \caption{Dataset sizes. Subscripts $A$ and $S$
    indicate article and sentence granularity, respectively.}
    \label{dataset}
    {%
        \resizebox{0.45\textwidth}{!}{
        \begin{tabular}{|c|cc|cc|c|}
        \hline
         &\multicolumn{2}{c|}{Speeches} &\multicolumn{2}{c|}{News} &Tweets\\
        Dataset & \spa & \sps & \hnda & \hnds & \ira\\
        \hline
        Size    &288 &24,934 &7,798 &7,876 &17,926 \\
        \hline
        \end{tabular}
        }
	}
\end{table}

\section{Methods}
In this section, we first introduce commonly used methods for propaganda detection, and subsequently introduce our proposed pairwise ranking model that aims at enhancing the effectiveness in the cross-domain setting. 
 
We investigate three widely used methods as propaganda classifiers:
logistic regression (\textbf{LR}) \cite{hosmer2013applied}, support vector machines (\textbf{SVM}) \cite{cortes1995support} and 
bidirectional long short-term memory (\textbf{LSTM}) \cite{graves2005framewise} neural networks. 
In addition, we devise an enhanced form of LSTM-based network
that incorporates pairwise ranking information and
is equipped with a specifically designed loss function.
We denote this method as \textbf{LSTMR}, and present it in the 
Subsection \ref{sec:model}.

\subsection{Feature-based Models}
\lr~and \svm~are feature-driven learners.
We consider various informative features on the language 
characteristics of
articles and sentences:
\squishlist
\item word counts (\textbf{WC}), 
\item word level TF-IDF scores (\textbf{TI-W}), 
\item N-gram level TF-IDF scores for $N$=2 or 3 (\textbf{TI-G}), 
\item occurrence statistics of word categories, such as first-person pronouns or negative-emotion words,
from the widely used Linguistic Inquiry and Word Count (\textbf{LIWC}) dictionary \cite{pennebaker2015development},
\item the combination of all the features (\textbf{ALL}).
\squishend

\subsection{LSTM Baseline Model}

Unlike LR and SVM, the LSTM-based methods do not rely on feature modeling.
Instead they can use pre-trained word embeddings to capture the text structure and semantic features.
We adopt a basic LSTM classifier as a baseline.
The structure of the LSTM baseline model is illustrated in Figure \ref{lstm-rank}. On top of the bi-LSTM layers, the networks have a dense layer
and a final sigmoid function to learn scores and yield classifier labels. 
The cross-entropy loss function is applied for the \lstm\ classifier.

\subsection{Pairwise Ranking Model}
\label{sec:model}

\begin{figure}[ht!]
    \centering
    \includegraphics[width=1\columnwidth]{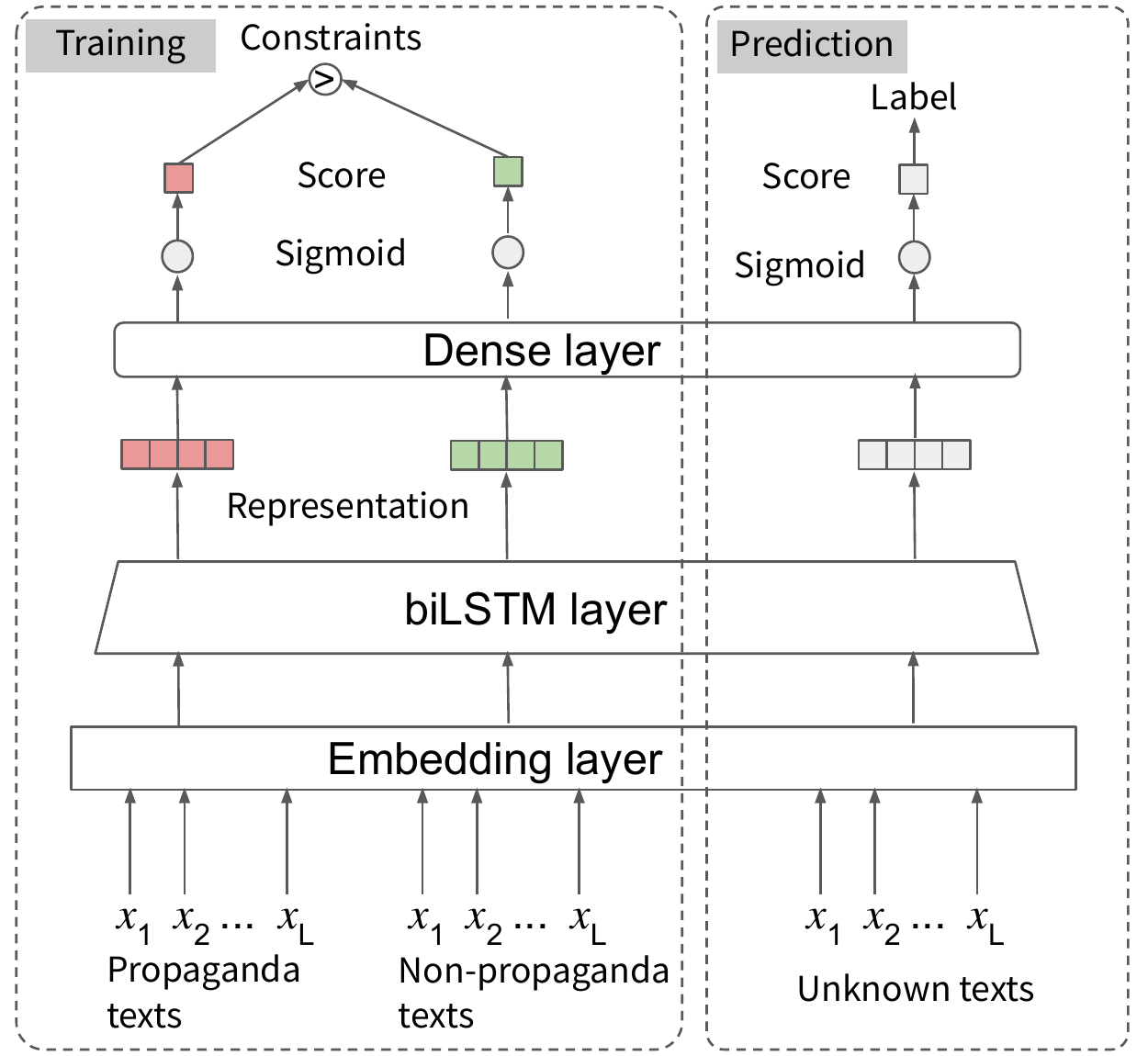}
	\caption{Pairwise propaganda ranking model \lstmr}
	\label{lstm-rank} 
\end{figure}

\pheadA{Model Architecture} Figure~\ref{lstm-rank} depicts our pairwise propaganda ranking model \lstmr. 
The motivation for this design is that training labels may be
subjective and noisy, especially but not only in a cross-domain setting
where training and test data do not come from the same distribution.
Traditional supervised learning models strictly distinguish the samples according to their labels, which leads to models over-fitting the training set, while their cross-domain generalizability diminishes. 
Supervised learning typically interprets the labels of data samples as a
strict ground truth. In certain settings, this can lead to over-fitting
to the training data distribution while hampering the ability for
cross-domain learning.
The \lstmr\ model aims to mitigate this over-fitting and enhance cross-domain applicability by relaxing constraints from strict labeling to rankings. 
Unavoidably, the in-domain performance may decrease. 
As Figure~\ref{lstm-rank} shows, the model operates in two phases:
\begin{enumerate}
    \item The first phase is to train the model based on the ranked
    pairs of data samples, where one is considered more propagandistic
    than the other. The bi-LSTM and dense layers serve to learn numeric scores
    for each of the two data points in each pair. 
    Then the two scores are constrained by a rule that the more propagandistic sample is supposed to obtain a higher score than the other sample that it is paired up with. We can think of the former as a positive sample and the latter as a negative sample, in terms of a binary classifier.
    This stage results in an improved scoring function, with awareness of the constraints.
    \item The second phase addresses the scoring of previously unseen data points with the trained \lstmr\ model receiving a single input text instead of a pair. The label is then determined according to the constraints-based scoring function.
\end{enumerate}

\pheadA{Pairwise Constraints} The constraints are instantiated using the following loss functions, where $\mathbf{y}$ is the ground-truth label ($0$ and $1$ corresponding to the less propagandistic and the more propagandistic sample, respectively),  $\hat{\mathbf{y}}$ is the predicted score, and $\mathbb{I}(\cdot)$ is the indicator function with output 1(constraints are satisfied) or 0:
\begin{itemize}
    \item Logistic loss (\textbf{LOG}) \cite{pasumarthi2019tf}:
    \begin{equation}
        \label{formula:logisticloss}
        \begin{aligned}
        \ell (\mathbf{y}, \hat{\mathbf{y}}) &= \sum_{j=1}^{n}  \sum_{k=1}^{n} \mathbb{I}(y_j>y_k) \cdot\\ &\log(1+\exp(\hat{y}_k-\hat{y}_j))
        \end{aligned}
        \end{equation}
    \item Linear loss (\textbf{LIN}) \LQ{\cite{hanselowski2018ukp}}:
        \begin{equation}
        \label{formula:linearloss}
        \begin{aligned}
        \ell (\mathbf{y}, \hat{\mathbf{y}})= \sum_{j=1}^{n}  \sum_{k=1}^{n} \mathbb{I}(y_j>y_k)\,(1+\hat{y}_k-\hat{y}_j)
        \end{aligned}
        \end{equation}
    \item Threshold loss (\textbf{THR}), where $\theta$ is the threshold to control the intensity of the constraints:
        \begin{equation}
        \label{formula:thr}
        \begin{aligned}
        \ell (\mathbf{y}, \hat{\mathbf{y}}) &= \sum_{j=1}^{n}  \sum_{k=1}^{n}  \mathbb{I}(y_j>y_k) \cdot\\ &\mathbb{I}\left(\frac{1+\hat{y}_k-\hat{y}_j}{2}>\theta\right) \frac{1+\hat{y}_k-\hat{y}_j}{2}
        \end{aligned}
        \end{equation}
    \item Counting loss (\textbf{COU}) to count the discordant pairs:
        \begin{equation}
        \label{formula:cou}
        \begin{aligned}
        \ell (\mathbf{y}, \hat{\mathbf{y}})= \sum_{j=1}^{n}  \sum_{k=1}^{n} \mathbb{I}(y_j>y_k)\mathbb{I}(\hat{y}_k>\hat{y}_j)
        \end{aligned}
        \end{equation}
\end{itemize}
Among the above loss functions, the {\bf threshold loss} is designed to further relax the constraints by explicitly disregarding those pairs with confidence below the threshold.
We compare the model performance using different loss functions in Section \ref{sec:ana}. In the subsequent evaluation experiments we adopt \thr\ loss function for \lstmr\ by reason of its better performance. 

\pheadA{Sampling Method} We apply different methods for 
constructing and sampling the ranked pairs.
We consider data from all three source types: news, tweets and speeches.
Whenever we have 0/1-labeled data points, we randomly combine 
a positive sample (1) and a negative sample (0) into an ordered pair
for training the \lstmr\ model. This is the case for news and tweets,
as these datasets have been manually annotated.
When we have merely relatively ordered data points where some are
more propagandistic than others, we sample such pairs. This is the
case for the speeches, where we assume that Goebbels is more
propagandistic than Churchill and the same ranking holds for
Trump versus Obama.

For news and tweets, we need to cope with skewed label distributions:
way more negative than positive samples.
As we consider cross-domain learning where the eventual test set
has an a-priori unknown distribution that could be fairly different
from the prior label distribution at training time,
we generally re-balance all datasets. We consider two strategies to this end.
The first is to under-sample the entire dataset, and the second is to over-sample the smaller class. 
Note that the over-sampling is per data point, so we still construct 
new pairs of ranked samples (i.e., there are no duplicate pairs).

\begin{table*}[!t]
\setlength{\tabcolsep}{1pt}
	\centering
    \caption{F1 score (\textperthousand) for in-domain classification with different methods, features and datasets. $PN$ stands for proper noun.}
    \label{tab:indomain}
    {%
        \resizebox{1\textwidth}{!}{
        \begin{tabular}{c|c|ccccc|ccccc|ccccc|ccccc|ccccc}
        \toprule
       \multirow{2}{*}{PN} &\multirow{2}{*}{Method}  &\multicolumn{5}{c|}{\spa} &\multicolumn{5}{c|}{\sps} &\multicolumn{5}{c|}{\ira} &\multicolumn{5}{c|}{\hnda} &\multicolumn{5}{c}{\hnds}\\
        & & WC &TI-W &TI-G &LIWC &ALL & WC &TI-W &TI-G &LIWC &ALL & WC &TI-W &TI-G &LIWC &ALL & WC &TI-W &TI-G &LIWC &ALL & WC &TI-W &TI-G &LIWC &ALL\\
\midrule
\multirow{4}{*}{YES}
&LR   &964 &971 &964 &903 &964 &844 &845 &795 &629 &\B{865} &813 &810 &651 &674 &\B{817} &914 &926 &895 &810 &\B{930} &665 &674 &621 &609 &677 \\
&SVM  &960 &982 &960 &900 &\B{982} &837 &838 &783 &630 &846 &795 &794 &638 &676 &795 &907 &915 &875 &812 &921 &647 &645 &598 &608 &657 \\
& LSTM &\multicolumn{5}{c|}{973} &\multicolumn{5}{c|}{747} &\multicolumn{5}{c|}{807} &\multicolumn{5}{c|}{866} &\multicolumn{5}{c}{715}\\
&  LSTMR &\multicolumn{5}{c|}{908} &\multicolumn{5}{c|}{751} &\multicolumn{5}{c|}{789} &\multicolumn{5}{c|}{856} &\multicolumn{5}{c}{\B{718}}\\
\midrule
\multirow{2}{*}{NO}
&LR   &960 &971 &960 &892 &964 &823 &820 &778 &621 &843 &755 &757 &665 &662 &762 &895 &905 &877 &811 &915 &646 &660 &615 &600 &659 \\
&SVM  &953 &982 &960 &900 &975 &816 &814 &766 &621 &820 &741 &739 &652 &661 &735 &882 &893 &853 &813 &907 &632 &639 &594 &595 &640 \\
        \bottomrule
        \end{tabular}
        }
	}
\end{table*}

\section{Experimental Evaluation}

\begin{table}[!ht]
\setlength{\tabcolsep}{1pt}
	\centering
    \caption{Precision (P), recall (R) and F1 (F) score (\textperthousand) on cross-domain classification with the features. Results are \colorbox{cadetgrey}{highlighted} when \lstmr~is better than \lstm. }
    \label{tab:cross}
    {%
        \resizebox{0.5\textwidth}{!}{
        \begin{tabular}{c|c|ccc|ccc|ccc|ccc|ccc}
        \toprule
       & Test  &\multicolumn{3}{c|}{\spa} &\multicolumn{3}{c|}{\sps} &\multicolumn{3}{c|}{\ira} &\multicolumn{3}{c|}{\hnda} &\multicolumn{3}{c}{\hnds}\\
        \cline{1-2}
        Train & Method &P &R &F &P &R &F &P &R &F &P &R &F &P &R &F\\
\midrule
\multirow{4}{*}{\spa}
&LR     &- &- &- &949 &084 &154 &451 &009 &017 &900 &005 &009 &696 &012 &024 \\
&SVM    &- &- &- &887 &181 &\B{301} &532 &056 &102 &771 &014 &027 &553 &055 &101 \\
\cline{2-17}
&LSTM   &- &- &- &639 &113 &191 &312 &118 &\B{172} &279 &190 &226 &450 &110 &\B{177} \\
&LSTMR  &- &- &- &632 &064 &115 &\cellcolor{cadetgrey}354 &067 &113 &\cellcolor{cadetgrey}300 &\cellcolor{cadetgrey}233 &\cellcolor{cadetgrey}\B{263} &430 &098 &159 \\
\midrule
\multirow{4}{*}{\sps}
&LR     &1000 &667 &800 &- &- &- &508 &648 &570 &333 &005 &010 &459 &367 &408 \\
&SVM    &1000 &778 &\B{875} &- &- &- &507 &630 &562 &440 &017 &033 &472 &400 &433 \\
\cline{2-17}
&LSTM   &886 &486 &628 &- &- &- &484 &889 &\B{627} &452 &018 &035 &485 &657 &\B{559} \\
&LSTMR  &\cellcolor{cadetgrey}980 &\cellcolor{cadetgrey}667 &\cellcolor{cadetgrey}793 &- &- &- &446 &587 &507 &\cellcolor{cadetgrey}475 &\cellcolor{cadetgrey}034 &\cellcolor{cadetgrey}\B{064} &\cellcolor{cadetgrey}490 &471 &480 \\
\midrule
\multirow{4}{*}{\ira}
&LR     &727 &056 &103 &537 &671 &596 &- &- &- &665 &192 &298 &524 &681 &592 \\
&SVM    &583 &049 &090 &524 &619 &568 &- &- &- &650 &246 &357 &533 &658 &589 \\
\cline{2-17}
&LSTM   &267 &306 &285 &517 &981 &\B{677} &- &- &- &537 &770 &633 &496 &961 &654 \\
&LSTMR  &\cellcolor{cadetgrey}356 &\cellcolor{cadetgrey}326 &\cellcolor{cadetgrey}\B{341} &500 &754 &601 &- &- &- &\cellcolor{cadetgrey}646 &669 &\cellcolor{cadetgrey}\B{658} &\cellcolor{cadetgrey}534 &855 &\cellcolor{cadetgrey}\B{657} \\ 
\midrule
\multirow{4}{*}{\hnda}
&LR     &510 &889 &648 &481 &556 &516 &571 &454 &506 &- &- &- &564 &685 &619 \\ 
&SVM    &500 &757 &602 &487 &572 &526 &554 &527 &540 &- &- &- &557 &702 &621 \\
\cline{2-17}
&LSTM   &516 &986 &\B{678} &362 &264 &305 &532 &197 &287 &- &- &- &616 &566 &590 \\
&LSTMR  &515 &972 &673 &\cellcolor{cadetgrey}489 &\cellcolor{cadetgrey}913 &\cellcolor{cadetgrey}\B{637} &\cellcolor{cadetgrey}532 &\cellcolor{cadetgrey}831 &\cellcolor{cadetgrey}\B{649} &- &- &- &530 &\cellcolor{cadetgrey}951 &\cellcolor{cadetgrey}\B{680}  \\
\midrule
\multirow{4}{*}{\hnds}
&LR     &503 &1000 &670 &460 &474 &467 &571 &393 &466 &514 &980 &674 &- &- &- \\
&SVM    &507 &1000 &673 &474 &486 &480 &546 &431 &481 &534 &964 &\B{688} &- &- &- \\
\cline{2-17}
&LSTM   &505 &1000 &671 &470 &750 &578 &559 &632 &593 &503 &\B{998} &669 &- &- &- \\
&LSTMR  &\cellcolor{cadetgrey}511 &\cellcolor{cadetgrey}1000 &\cellcolor{cadetgrey}\B{676} &\cellcolor{cadetgrey}476 &\cellcolor{cadetgrey}762 &\cellcolor{cadetgrey}\B{586} &\cellcolor{cadetgrey}570 &\cellcolor{cadetgrey}723 &\cellcolor{cadetgrey}\B{637} &\cellcolor{cadetgrey}511 &994 &\cellcolor{cadetgrey}675 &- &- &- \\
        \bottomrule
        \end{tabular}
        }
	}
\end{table}

\begin{table}[!ht]
\setlength{\tabcolsep}{1pt}
	\centering
    \caption{F1 score (\textperthousand) for cross-domain classification with proper nouns removed.}
    \label{tab:cross-pn}
    {%
        \resizebox{0.5\textwidth}{!}{
        \begin{tabular}{c|c|ccc|ccc|ccc|ccc|ccc}
        \toprule
       & Test  &\multicolumn{3}{c|}{\spa} &\multicolumn{3}{c|}{\sps} &\multicolumn{3}{c|}{\ira} &\multicolumn{3}{c|}{\hnda} &\multicolumn{3}{c}{\hnds}\\
        \cline{1-2}
        Train & Method &P &R &F &P &R &F &P &R &F &P &R &F &P &R &F\\
\midrule
\multirow{2}{*}{\spa}
&LR     &- &- &- &967 &061 &115 &491 &006 &013 &938 &004 &008 &688 &008 &017 \\
&SVM    &- &- &- &925 &098 &\B{178} &546 &026 &\B{050} &778 &007 &\B{014} &535 &023 &\B{045} \\
\midrule
\multirow{2}{*}{\sps}
&LR     &980 &347 &513 &- &- &- &494 &581 &\B{534} &207 &002 &003 &449 &347 &391 \\
&SVM    &987 &535 &\B{694} &- &- &- &497 &540 &518 &370 &008 &\B{015} &477 &371 &\B{417} \\
\midrule
\multirow{2}{*}{\ira}
&LR     &462 &042 &\B{076} &537 &704 &\B{609} &- &- &- &564 &215 &311 &520 &709 &\B{600} \\
&SVM    &333 &028 &051 &525 &641 &577 &- &- &- &566 &252 &\B{349} &525 &670 &589 \\
\midrule
\multirow{2}{*}{\hnda}
&LR     &518 &785 &\B{624} &486 &422 &452 &569 &341 &427 &- &- &- &576 &560 &568 \\
&SVM    &486 &611 &542 &490 &449 &\B{469} &532 &408 &\B{462} &- &- &- &564 &582 &\B{573} \\
\midrule
\multirow{2}{*}{\hnds}
&LR     &503 &1000 &670 &476 &491 &483 &539 &379 &445 &511 &982 &672 &- &- &- \\
&SVM    &505 &1000 &\B{671} &494 &506 &\B{500} &522 &432 &\B{473} &534 &961 &\B{687} &- &- &- \\
        \bottomrule
        \end{tabular}
        }
	}
\end{table}

\pheadA{Setup}
The feature dimensions of \wc, \tiw, \tig\ are truncated to their
respective top 5,000, and we use LIWC 2015 \cite{pennebaker2015development} with 76 categories including {\em first person singular}, 
{\em negation}, {\em sexual}, {\em swear}, etc.
Optionally, we eliminate proper nouns (\textbf{PN}), as tagged using the NLTK toolkit \cite{bird2009natural}.
This is to reduce the influence of particular names and their respective
topics (e.g., ``Iran''). The rationale is that
texts could be about the very same topic but differ in their degree
of propaganda.

\LQ{For \lstm\ and \lstmr, we use the pre-trained 100-dimensional embeddings from GloVe \citep{pennington2014glove}, 128 bi-LSTM cells and 200$\times$50$\times$1 dense layers. The embeddings are frozen during training. For \lstmr, the \thr~loss with a threshold of $0.4$ and complete sampling (neither under- nor over-sampling) is adopted.} Note that \lstmr~is a ranking model rather than a classifier. As the ranking scores are comparable and range from 0 to 1, we empirically consider $\geq 0.5$ as a threshold to classify test samples as positive or negative.

\LQ{For in-domain learning, where training and test data are from the same source,
we adopt 5-fold cross validation. Hyper-parameters are tuned by this setting,
and then kept fixed for cross-domain experiments. 
Note that in the cross-domain case, cross-validation is impossible 
as the target-domain data is treated completely unseen.}
We report on precision (\textbf{P}), recall (\textbf{R}) and F1 (\textbf{F}) scores, with propaganda
being the positive class.
\pheadA{In-Domain Performance} 
The results are shown in Table~\ref{tab:indomain}.
Overall, we achieve acceptable in-domain performance on most datasets (over 80\%) except \hnds~(around 65\%). 
Results for article-level classification (\spa~and \hnda) 
are better than for the sparser, and perhaps noisier,
sentence-level data (\sps, \ira~and \hnds). 
\LQ{We observe a larger drop for \hnds\ than \sps\ regarding the \lr\ and \svm\ methods but not for the \lstm\ and \lstmr\ methods. This suggests a greater robustness of semantic features compared to linguistic features.}

For the \lr~and \svm~ methods, the \wc, \tiw\ and \tig\ features work better than the \liwc\ features. However, the combination of all four feature groups yields the best results. 
\LQ{When omitting proper nouns, performance drops, which indicates that the models also learn some topic features instead of propaganda itself.}
\LQ{However, the \liwc\ representations are not affected much when excluding proper nouns due to their focus on word categories.}

The \lstm~and \lstmr\ methods perform slightly worse than \lr\ and \svm\ on all datasets except \hnds. \LQ{Due to the influence of the pairwise ranking and the loss function, \lstmr\ performs slightly worse than \lstm\ on some datasets. As a trade-off, it gets better cross-domain performance (see Table \ref{tab:cross}).}
Overall, feature-based learning works best for the 
standard case of in-domain classification.

\pheadA{Cross-Domain Performance} 
The cross-domain results are given in Table~\ref{tab:cross}. As expected, cross-domain classification is much more challenging than in-domain classification. Still, many cross-domain settings show reasonably good performance. 
The best cross-domain results are observed for the classifier trained on \hnda\ and tested on \hnds. The most balanced precision--recall performance is obtained by the classifier trained on \ira\ and tested on \hnds. 
Overall, \lstmr\ compares favorably against other methods, especially on \ira, \hnda\ and \hnds.

\pheadA{Combining Datasets for Training} 
As a final configuration, we combine several datasets, excluding \hnds, into a single training set and test on \hnds. The results are given in Table~\ref{tab:cross-amount}. 
As we see, the larger amount of training data does not help the classifiers
to improve their cross-domain performance. On the contrary, the F1 scores
slightly drop when more data is combined. The likely reason is the high
variability in the underlying features and data characteristics that
comes from such highly heterogeneous sources.
These settings require further research.

\begin{table}[!ht]
\setlength{\tabcolsep}{3pt}
	\centering
    \caption{Cross-domain performance 
    (\textperthousand) 
    of classifiers trained on combined datasets and tested on \hnds.}
    \label{tab:cross-amount}
    {%
        \resizebox{0.5\textwidth}{!}{
        \begin{tabular}{c|ccc|ccc|ccc|ccc}
        \toprule
       \multirow{2}{*}{Method}  &\multicolumn{3}{c|}{\spa} &\multicolumn{3}{c|}{+\sps} &\multicolumn{3}{c|}{++\ira} &\multicolumn{3}{c}{+++\hnda}\\
         &P &R &F &P &R &F &P &R &F &P &R &F\\
\midrule
LR     &696 &012 &024 &447 &327 &377 &503 &446 &473 &509 &580 &\B{542}\\
SVM    &553 &055 &101 &485 &363 &415 &508 &475 &491 &507 &554 &529\\
LSTM   &450 &110 &\B{177}  &478 &594 &\B{530}  &496 &966 &\B{655} &462 &639 &536\\
LSTMR  &430 &098 &159  &477 &416 &445 &504 &361 &421 &508 &458 &482 \\
        \bottomrule
        \end{tabular}
        }
	}
\end{table}

\section{Analysis}
\label{sec:ana}

\pheadA{Datasets and Granularity}
The best cross-domain results are obtained for training on news and applying the learned models to speeches or tweets. This suggests that high-quality
training labels are still crucial. 
\LQ{Performance for articles is better than for sentences,
underlining the difficulty of dealing with very short texts without
placing them into their full context. From a model perspective, this makes features such as TF-IDF sparser without abundant signals to aid in the detection.}

\pheadA{Proper Nouns} Table~\ref{tab:cross-pn} shows cross-domain results when excluding proper nouns. 
We observe a notable drop in performance. This suggests that 
focusing on language
alone is not sufficient for detecting propaganda.
The entities to which news, tweets or speeches refer are important
as topical context, and cannot be easily left out.
\GW{This finding seems to contradict recent experimental work on fact checking \cite{DBLP:conf/emnlp/SuntwalPSS19}, where
de-lexicalization was found useful.
A major reason is that our cross-domain setting not just switches between data sources, but involves a transfer to a different style of source, like news vs.\ speeches. 
Moreover, our negative finding underlines the additional complexity 
of propaganda detection, 
calling for more research on the role of
proper nouns.
}%

\pheadA{Feature Groups} We conduct experiments with classifiers
learned with single feature groups, using \hnds\ for training.
Table~\ref{tab:cross-feature} gives the cross-domain results.
Comparing these against the \all\ feature performance in Tables~\ref{tab:indomain} and \ref{tab:cross-pn}, 
we observe that the combination of all four features has a positive effect
for in-domain classification but does not work so well for cross-domain learning. However, there is no single feature group that dominates all others. 
This suggests that different datasets exhibit different kinds of linguistic cues,
underlining again the big challenge in cross-domain classification.

\begin{table}[!ht]
\setlength{\tabcolsep}{2pt}
	\centering
    \caption{Cross-domain performance (F1 score, \textperthousand)  with training on \hnds, using different features. `1', `2', `3' and `4' stand for \wc, \tiw, \tig\ and \liwc.}
    \label{tab:cross-feature}
    {%
        \resizebox{0.5\textwidth}{!}{
        \begin{tabular}{c|cccc|cccc|cccc|cccc}
        \toprule
       \multirow{2}{*}{Method}  &\multicolumn{4}{c|}{\spa} &\multicolumn{4}{c|}{\sps} &\multicolumn{4}{c|}{\ira} &\multicolumn{4}{c}{\hnds}\\
         & 1 &2 &3 &4& 1 &2 &3 &4& 1 &2 &3 &4& 1 &2 &3 &4\\
\midrule
LR      &673 &673 &671 &387&397 &458 &452 &491 &422 &487 &418 &435&672 &676 &623 &603\\
SVM     &674 &673 &\B{678} &390&428 &467 &458 &\B{493} &459 &\B{498} &416 &443&687 &\B{688} &662 &596\\
        \bottomrule
        \end{tabular}
        }
	}
\end{table}

\pheadA{Informative Language Features} To analyze to what extent
word-level features can yield interpretable cues that characterize
propaganda in different kinds of sources,
we trained the  \lr\ classifer on the \hnds\ data with \tiw\ and \liwc\ features, %
excluding proper nouns. 
We observed the following most distinctive words, characteristic for propaganda:
\squishlist
\item[$~$] %
``advantage'', ``absolutely'', ``neo'', ``political'', ``attacks'', ``american'', ``shocking'', ``influential'', ``impossible'', ``lies'', ``devastating'', ``hell'', ``administration'', ``autonomous'', ``ridiculous'' 
\squishend
The most salient \liwc\ features (word categories) were:
\squishlist
\item[$~$] %
\emph{affect}, \emph{neg-emo}, \emph{anger}, \emph{swear}, \emph{certain}
\squishend

This suggests that exaggerations (e.g., ``absolutely'') and 
negative emotions (e.g., ``lies'' or ``devastating'') play a key role
in manipulating the audience.
As for \liwc\ features, words that express negative emotions 
are typical for propaganda, as well as being strongly self-confident.

\begin{figure}[htp]
    \centering
    \includegraphics[width=0.95\columnwidth]{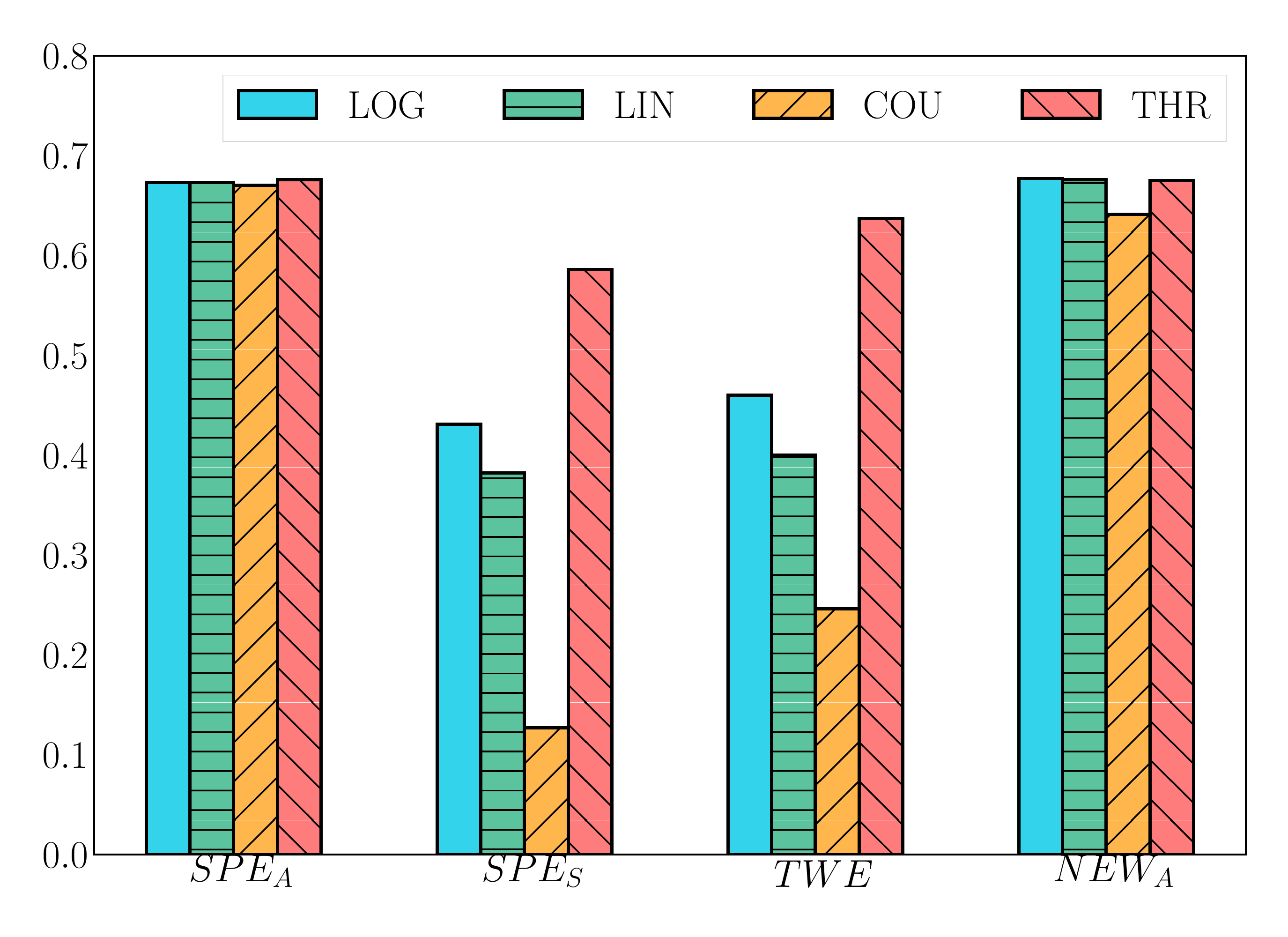}
	\caption{F1 of  \lstmr\ trained on \hnds\ and tested on other datasets, with different loss functions.}
	\label{exp-loss} 
\end{figure}

\pheadA{LSTMR Loss Functions}
For our \lstmr\ model, the selection of the loss function 
has a notable influence on performance. 
The previous experiments all used \thr\ as loss function.
Training on \hnds\ and with cross-domain test data, 
Figure~\ref{exp-loss} compares the effectiveness of all four loss functions that we studied. 
We notice that the specifically designed \thr\ loss function outperforms all
alternatives, especially on
the \sps\ and \ira\ datasets. 
This observation is in line with our design goal that \thr\ 
helps the model training to discount ranked pairs of low confidence. 
We sacrifice in-domain performance to some extent,
but improve the cross-domain classifier.

\begin{figure}[htp]
    \centering
    \includegraphics[width=1\columnwidth]{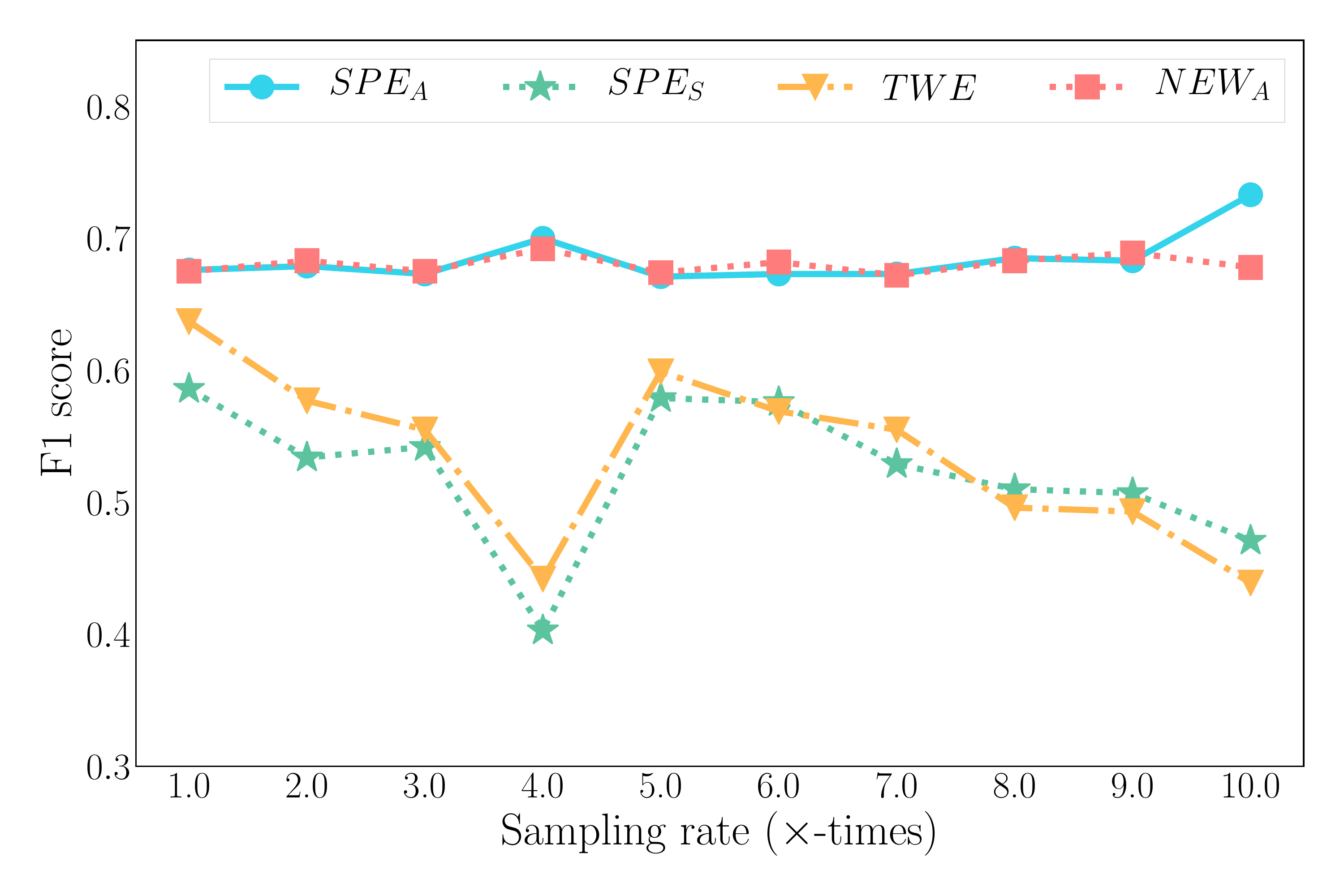}
	\caption{F1 of \lstmr\ trained on \hnds\ and tested on other data, with varying rate of over-sampling. 
	}
	\label{sample-over} 
\end{figure}
\begin{figure}[htp]
    \centering
    \includegraphics[width=1\columnwidth]{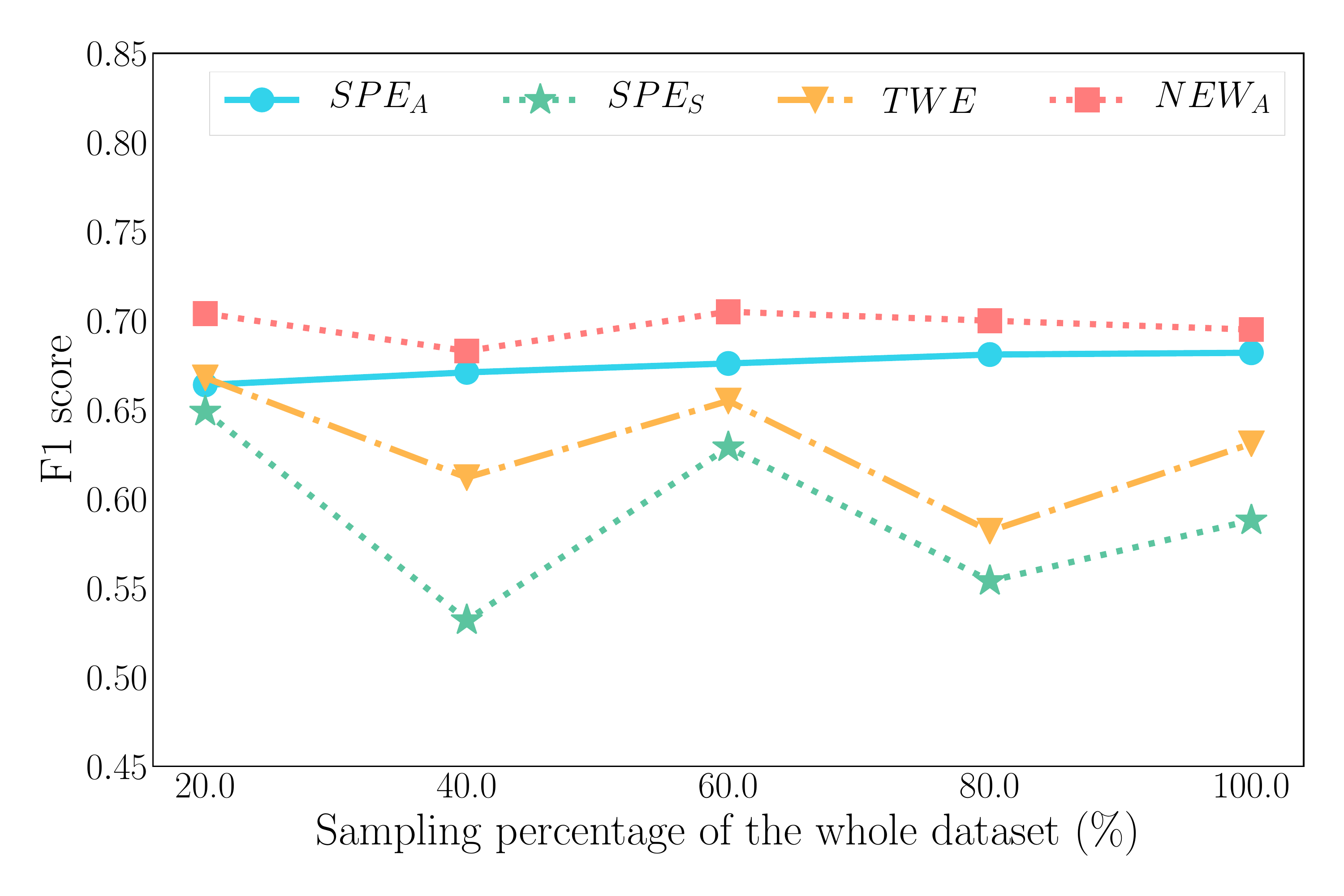}
\caption{F1 of \lstmr\ trained on \hnds\ and tested on other data, with varying percentage of under-sampling. 
	}
	\label{sample-under} 
\end{figure}

\pheadA{Sampling Strategies}
The training of \lstmr\ requires ranked pairs of samples. 
Over-sampling constructs these pairs by repeatedly drawing from
the positive and negative classes.
The effect of the amount of over-sampling is plotted in Figure~\ref{sample-over}, %
for the setting with training on \hnds\ and cross-domain testing on the other datasets. The results show the limitations of this sampling strategy:
there is hardly any increase in performance and even a drop for some cases.
This calls for further research towards overcoming training bottlenecks.

For the under-sampling strategy, Figure~\ref{sample-under} shows results
(for the same train-test setting) with varying fractions of training samples. 
On the \spa\ data, the performance increases with more training samples, as expected. However, on other test datasets, performance remains unchanged
or drops or fluctuates.
This points out the limitations of cross-domain classification.
Our approach counters the training bottleneck to a good degree,
but there is quite some room for improvement
and the need for further research.

\section{Conclusion and Future Work}
\balance

Although propaganda has become a pervasive challenge in online media,  previous work has mostly treated it 
as a variation
of fake news, or considered unrealistic settings where the test distribution precisely matches the training data distribution.

In this paper, we present 
\GW{a first and preliminary}
analysis of the problem of propaganda detection in cross-domain learning settings. This encompasses several novel aspects, ranging from data collection methods, feature computation, designing different classifiers, and the corresponding analysis. 
\GW{We tap into a previously unexplored content source: speeches by politicans who are known
for different levels of propaganda, using them as collective and relative signals.}
On the methodology side, we devise a pairwise ranking method with customized loss functions to improve the classification. The experimental results demonstrate the effectiveness of this method. Furthermore, we conduct a series of experiments to explore the most salient factors for cross-domain generalizability of propaganda detection learning. The observations and analysis reveal insightful patterns and lessons for building  more general propaganda detectors.

\GW{As our datasets are still fairly small, 
our findings are of preliminary nature
and our methodology is subject to ongoing research.
We believe that cross-domain learning is a crucial asset for
the important topic of propaganda detection, and hope that
our initial results are useful for further research along these lines.}

\section*{Acknowledgments}
The work is partly supported by ``The Fundamental Research Funds of Shandong University''. 
Gerhard Wei\-kum's work is partly supported by the
ERC Synergy Grant 610150 (imPACT). 
Any opinions, findings and conclusions, or recommendations expressed in this material are those of the authors and do not necessarily reflect the views of the funding agencies.
\bibliography{acl2019}
\bibliographystyle{acl_natbib}
\end{document}